\documentclass[letterpaper, 10 pt, conference]{ieeeconf}  % Comment this line out if you need a4paper

\IEEEoverridecommandlockouts                              % This command is only needed if 
\usepackage[%dvipdfm,
bookmarks=true,
bookmarksnumbered=true,
bookmarkstype=toc, %
pdfauthor={Liao Wu},%
pdftitle={RAL},%
pagebackref=false,
colorlinks=true,
citecolor=red,
linkcolor=blue,
anchorcolor=red]{hyperref}

\usepackage{graphicx} % for pdf, bitmapped graphics files
\usepackage{epsfig} % for postscript graphics files
\usepackage{mathptmx} % assumes new font selection scheme installed
\usepackage{times} % assumes new font selection scheme installed
\usepackage{amsmath} % assumes amsmath package installed
\usepackage{amssymb}  % assumes amsmath package installed
\usepackage{booktabs} %for table
\usepackage{multirow}
\usepackage{cite}
\usepackage{xcolor}
\usepackage{bm}
\usepackage{tabularx}
\usepackage{hyperref}

\title{\LARGE \bf
Model-less Active Compliance for Continuum Robots using Recurrent Neural Networks}

\author{David Jakes, Zongyuan Ge, and Liao~Wu,~\IEEEmembership{Member,~IEEE}% <-this % stops a space
	\thanks{D. Jakes is with the School of Electrical Engineering and Computer Science, Queensland University of Technology, Brisbane, Australia.}
	\thanks{Z. Ge is with eResearch Centre, Nvidia AI technology centre, Airdoc Research, Monash University, Melbourne, Australia.}
	\thanks{L. Wu is with the School of Mechanical and Manufacturing Engineering, University of New South Wales, Sydney, Australia. \tt\small dr.liao.wu@ieee.org}
	\thanks{A multimedia attachment demonstrating the experiments can be found at \href{https://youtu.be/y5iW1QgeTBg}{https://youtu.be/y5iW1QgeTBg}}}

\begin{document}

\maketitle
%\thispagestyle{empty}
%\pagestyle{empty}

%%%%%%%%%%%%%%%%%%%%%%%%%%%%%%%%%%%%%%%%%%%%%%%%%%%%%%%%%%%%%%%%%%%%%%%%%%%%%%%%
\begin{abstract}
Endowing continuum robots with compliance while it interacts with the internal environment of the human body is essential to prevent damage to the robot and the surrounding tissues.
Compared with passive compliance, active compliance has the advantages in terms of increasing the force transmission ability and improving safety with monitored force output.
Previous studies have demonstrated that active compliance can be achieved based on a complex model of the mechanics combined with a traditional machine learning technique such as a support vector machine.
This paper proposes a recurrent neural network (RNN) based approach that avoids the complexity of modeling while capturing nonlinear factors such as hysteresis, friction and delay of the electronics that are not easy to model.
The approach is tested on a 3-tendon single-segment continuum robot with force sensors on each cable.
Experiments are conducted to demonstrate that the continuum robot with an RNN based feed-forward controller is capable of responding to external forces quickly and entering an unknown environment compliantly.
\end{abstract}

%\begin{Keywords}
%Active Compliance, Continuum Robots, Recurrent Neural Networks, Model-less
%\end{Keywords}

%%%%%%%%%%%%%%%%%%%%%%%%%%%%%%%%%%%%%%%%%%%%%%%%%%%%%%%%%%%%%%%%%%%%%%%%%%%%%%%%
\section{Introduction}
Continuum robots have been identified as a class of robots that are ideal for minimally invasive surgery \cite{burgner2015continuum}. 
Applications of these robots span from neurosurgeries \cite{butler2012robotic},  trans-oral and trans-nasal procedures \cite{simaan2009design,wu2016development,yu2016development}, intracardiac procedures \cite{gosline2012percutaneous}, to orthopaedic surgeries \cite{paul2017prototype}. 
Compared to their discrete rigid counterparts, the continuum robots can be made very slim due to the novel actuation methods used that leverage the deformation of the material that constructs the body of the robot.
When used in surgical applications, this means the robot can be inserted into the human body through a small incision/orifice and a narrow passage, being minimally invasive to the patient.

Safety is one of the key issues to address in surgeries using continuum robots.
Endowing a robot with compliance while it interacts with the environment is essential to prevent damage to the robot and the surrounding tissues. 
Passive compliance can be realized by the nature of many mechanisms for the construction of continuum robots like the cable-driven mechanism \cite{gravagne2002manipulability}.
However, this approach limits the force transmission ability for tasks that are force demanding, such as tissue dissection and retraction, and increases the uncertainty of control due to the unmonitored flexibility while the robot is operating in the human body.

Recently, active compliant motion control for continuum robots has been studied \cite{goldman2011compliant,goldman2014compliant}.
By measuring joint level actuation forces and relating them to the generalized forces in the configuration space that are mapped from the external wrenches, it was demonstrated that the active compliance can be achieved without expensive sensors deployed along the distal part of the robot.
However, this was done based on a complex nonlinear model of the kinematics and mechanics of the structure.
In addition to the complexity in analysis and implementation, there are also many factors that cannot be easily modeled, such as the hysteresis of cables, backlash in transmission, friction, characteristics of the electronic components, etc.

Machine learning techniques could be used to bypass complex modeling and tackle those nonlinear factors.
In \cite{goldman2014compliant}, a support vector machine (SVM) was used to capture the friction and correct the uncertainties in the complex modeling.
However, the method still relied on elaborated analysis and modeling of the mechanics of the continuum robot.
We argue that a model-less compliant motion control strategy could be achieved in the machine learning framework by directly relating the actuation with the force measurement.
In addition, instead of using an SVM, the neural network (NN) based techniques, representing the-state-of-the-art, could produce better results in terms of capturing the nonlinear characteristics in the compliant motion control.
In particular, the recurrent neural network (RNN)~\cite{mikolov2010recurrent,mayer2008system} is a good tool to deal with the process-dependent factors such as hysteresis of cables and delay of electronic components.
%Examples of recurrent neural network for robotics applications can be seen \cite{mayer2008system}.

\begin{figure}[tb]
	\centering
	\includegraphics[width=8cm,keepaspectratio]{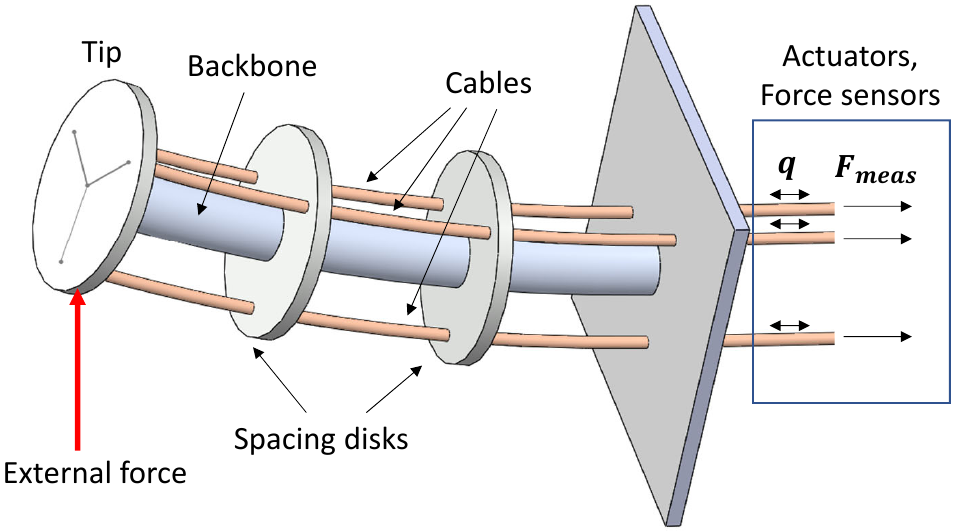}
	\caption{A 3-tendon single-segment continuum robot with force sensors on each cable. Active compliance allows the robot to move compliantly by adjusting the actuating cables when experiencing external forces along the segment.}
	\label{fig:tendon3}
\end{figure}

% Summary of method/contribution
This paper proposes a new method for compliant motion control of continuum robots by using an RNN.
Using a 3-tendon single-segment continuum robot with force sensors on each cable (Fig. \ref{fig:tendon3}), an RNN is trained to predict the cable forces produced by backbone stiffness, friction and other factors from the actuator control signals during unloaded motion. 
Actuator positions are supplied to the RNN producing a feed-forward prediction to a simple motion controller which achieves compliance by directly controlling actuators to minimize the difference between predicted and measured tension.
Comparison between RNN and other architectures is performed to show the superiority of RNN for this task. 
We validate the proposed approach on an experimental platform, demonstrating the effectiveness of a feed-forward RNN compliance controller which can bypass the complexity of nonlinear modeling and capture the effect of physical and electrical factors that are not modelable in continuum robots.

The rest of the paper is organized as follows.
Sec. II describes the problem of active compliance and introduces the role of RNN in the framework.
Sec. III gives a more detailed introduction to the RNN architecture used in this paper.
Based on the above, a compliant motion control algorithm is presented in Sec. IV.
Experimental validation of the method is elaborated in Sec. V to show the effectiveness.
The paper is discussed and concluded in Sec. VI and VII, respectively.

\section{Problem description}
%Describe the platform studied in this paper.
The design of continuum robots featuring an elastic backbone driven by three or more evenly-spaced tendons routed along the periphery of the tube is both prominent and well-validated \cite{walker2013continuous}. 
A key control requirement is that tendon tension remains within safe ranges as excessive tension may cause tendons to snap, while slack can invalidate modeling assumptions and sensor measurements causing backlash and potential structural damage. 
The analysis in this section is performed on a single-segment robot with three tendons. 
Force sensors are located at the connection point between tendons and actuators to measure individual tendon tension. 

Active compliance control requires the movement of actuators $\bm{q}$ in a way that minimizes and therefore complies with measured external forces $\bm{F_{ext}}$. 
Assuming backbone incompressibility, external forces applied to the tip of the robot are balanced by increased tension in the tendons providing the robot configuration holding force.
Ref \cite{goldman2011compliant,goldman2014compliant} have demonstrated that interaction forces at the tip and along the length of the backbone can be reconstructed from measurements of tendon forces. 

In addition to external forces, sensor measurements $\bm{F_{meas}}$ also include internal forces $\bm{F_{int}}$ caused by the robot's natural elastic deformation, gravity, friction and dynamics while in motion, i.e., \begin{equation}
\bm{F_{meas}} = \bm{F_{int}} + \bm{F_{ext}}.
\end{equation}

$\bm{F_{int}}$ has a fixed but unknown relationship to the actuator state space,
\begin{equation}\label{eq:f_int}
\bm{F_{int}} = f \left(\bm{q}, \Dot{\bm{q}}, \Ddot{\bm{q}},...\right).
\end{equation}
In continuum robots, these governing equations are complex and highly nonlinear incurring significant modeling and computational effort.

The most prevalent model for static forces in continuum robots is the Cosserat rod theory \cite{mahvash2011stiffness} using an elastic energy formulation to derive forces and curvature from backbone stiffness and cable lengths.
However, this model is purely static physics based and does not take effects accompanying dynamic motions into account.

%\color{red!60}
%--description of theory (I don't understand it well enough currently), model works well for static forces, but insufficient for motion--
%\color{black}
Actuating the cables for change in configuration involves acceleration, backbone momentum and the conversion of elastic to kinetic energy, producing forces which are translated across the tendons. Typically, these forces are ignored under small velocities with the assumption that small mass and velocity produces negligible momentum and the elastic restoration rate is sufficiently fast. 

Directional hysteresis, where transient forces depend on the direction of motion due to the microstructure of the elastic backbone, has been addressed using piece-wise curves selected based on the sign of the velocity \cite{goldman2011compliant}.

\begin{figure}[tb]
	\centering
	\includegraphics[width=6cm,keepaspectratio]{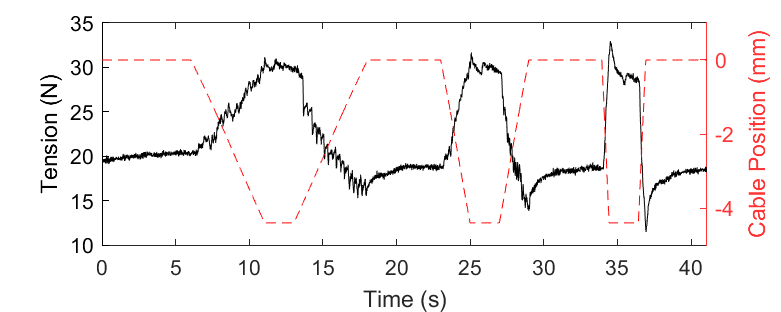}
	\caption{Cable tension throughout linear motion. The first, second, and third waveform corresponds to cable motion speed at 1mm/s, 2.5mm/s, and 10mm/s, respectively. }
	\label{fig:rate_statics}
\end{figure}

These deviations from the static model can be observed in Fig.~\ref{fig:rate_statics}. 
At slow speeds, the cable tension is dominated by static elastic stiffness with a small directional hysteresis component adding an offset when increasing versus decreasing tension. 
As movement speed increases, the deviation from the static model increases and an overshoot-with-restitution curve is observed accompanying sudden changes in velocity.
This is mainly due to effects such as friction, slip, and other uncertainties that are difficult to capture with traditional modeling. 

%Ref  \cite{goldman2011compliant} demonstrated a compliant controller with such a formulation as
% \begin{equation}\label{eq:elast_hyst}
% \bm{F_{int}} = elastic(\bm{x}) + hysteresis(sign(\bm{\dot{x}})).
% \end{equation}

In the past, machine learning has been used to augment kinematic models, achieving desired control stability/accuracy using feed-forward compensation for friction and other model uncertainties \cite{goldman2014compliant}. 
Given the complexity of kinematic modeling and partial application of machine learning techniques already in use, there is motivation for replacing the combination of techniques with a single machine learning algorithm producing feed-forward force estimates. 

\begin{figure}[tb]
	\centering
	\includegraphics[width=6cm,keepaspectratio]{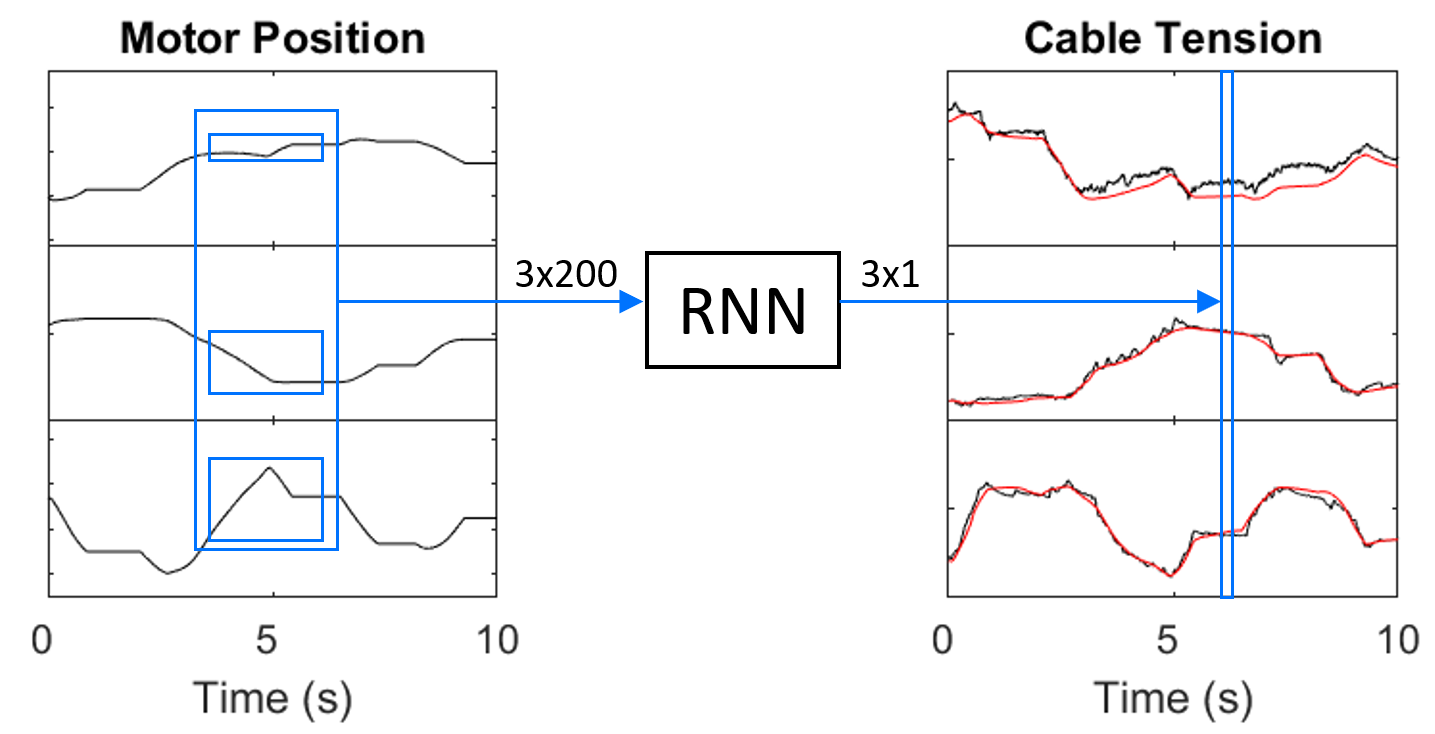}
	\caption{An RNN based tension predictor is supplied with $n=200$ discrete time points of control signal for each actuator, and predicts the tension in each cable at the next timestep $t+1$. Recorded tension training data (black) is shown with prediction tension (red).}
	\label{fig:nn_io}
\end{figure}

In this paper, we propose to use an RNN for the formulation of (\ref{eq:f_int}).
The input matrix is formed by concatenating the actuator control vector $\bm{q}$ over a period of recent discrete timesteps,
\begin{equation}
    Q = 
    \begin{bmatrix}
        \bm{q_{t}} & \bm{q_{t-\Delta t}} & ... & \bm{q_{t-n\Delta t}}
    \end{bmatrix},
\end{equation}
and the network learns the function,
\begin{equation}
\bm{F_{int}} = rnn(Q).
\end{equation}
The concept of using RNN here is to use the continuous control input signal to predict the measured tension resulting from electrical and physical statics and dynamics.
Fig.~\ref{fig:nn_io} shows how RNN is employed in this work for tension prediction.
A detailed description of RNN is presented in the next section.

% \color{red!60}
% --neural networks can model arbitary nonlinear functions--
% --RNN structure is suitable for task (see eqn 2)--
% --Hidden state can store velocity, acceleration, and other short-term memory properties such as hysteresis, delays and restitution--
% --Potentially mention forward difference constructions, or expanded form of RNN in a discrete time application--
%1st time derivative is just x_n - x_{n-1}
%Delays x_{n-\delta t/h}
%x_n -\textgreater x_n-200

% --continuous task, use the control input to predict the measured tension resulting from electrical and physical statics and dynamics, refer to Fig. \ref{fig:rate_statics} and Fig. \ref{fig:nn_io}--

\section{Recurrent Neural Network}
\color{black}
Since our aim is to characterize the high-order correlation between sensor measurements and actuator state space, we implement RNN with especially long short term memory (LSTM) neurons as recurrent neurons that can effectively model the long-term temporal dependency in a continuous signal space. RNN is a neural network that uses internal hidden states to model the dynamic behaviour of sequences with arbitrary nonlinear functions. 
In this task, the hidden states may store relevant long-term temporal dependencies of velocity, acceleration, and other short-term memory properties such as hysteresis, delays and restitution. The full formulation of a typical LSTM is defined as follows,
\begin{eqnarray}    
    i_{t} &=& \sigma(W_{q,i}\bm{q}_{t}+W_{h,i}h_{t-1}) \\ \nonumber
    f_{t} &=& \sigma(W_{q,f}\bm{q}_{t}+W_{h,f}h_{t-1}) \\ \nonumber
    o_{t} &=& \sigma(W_{q,o}\bm{q}_{t}+W_{h,o}h_{t-1}) \\ \nonumber
    g_{t} &=& \phi(W_{q,c}\bm{q}_{t}+W_{h,c}{h_{t-1}}) \\ \nonumber
    c_{t} &=& f_{t} \odot c_{t-1} + i_{t} \odot g_{t} \\ \nonumber
    h_{t} &=& o_{t} \odot \phi(c_{t})
\end{eqnarray}
\noindent In addition to RNN that has the hidden state $h$ with $N$ hidden units,  LSTM has five extra gates compared to an RNN: an input gate $i$ and input modulation gate $g$ to indicate the input information ratio; a forget gate $f$ and memory gate $c$ to control the memory of the current state; an output gate $o$ to control the level of output to the state.   
$\sigma$ is the \textit{sigmoid} nonlinearity which takes the input and output real-valued inputs to a $[0,1]$ range while $\phi$ is denoted as \textit{hyperbolic tangent} nonlinearity function similar to \textit{sigmoid} but with range of $[-1,1]$. 
These components make LSTM easier to optimize during training progress by back-propagating gradients further and thus enable the model to learn extremely long-term and complex dependency. 

%\subsection{RNN model details}
In this paper, we build the LSTM model by feeding input to the $n$ LSTM cells ($n=32,64$) followed by a 32 neurons fully connected linear layer. 
One important issue of training an RNN model is to determine the input sequential length. We set the sequence length of 100 and 200 to validate the effectiveness of the model with respect to the time factor.  
By combining all input signals from three motors into a single hidden state, the network learns a holistic representation of the model, allowing common features to be combined with different weights to produce each of the three outputs.
Training data is generated by recording tension measurements during unloaded motion using a rudimentary non-compliant control algorithm which traverses the available actuator space across a range of motion styles. More details about sample collection are described in Sec. \ref{sec:training}.

Training LSTM model can be achieved by using the mean squared error (MSE) loss on the three output of fully connected layer and employing a back-propagation algorithm with stochastic gradient descent (SGD). 
% To avoid overfitting issue, we attach a dropout layer with $p=0.2$ between the LSTM and linear layer. 
We employ a momentum rate of 0.9 and a learning rate of 0.005 for all experiments. 

% The input matrix is supplied to a stack of 64 LSTM cells producing a 64 element feature vector, followed by a fully connected linear layer with 3 outputs. By combining all input channels into a single hidden state, the network learns a holistic representation of the model, allowing common features to be combined with different weights to produce each of the 3 outputs.

% A dropout layer is added with $p=0.2$ between the LSTM and Linear layers to reduce overfitting. Training is performed offline with an MSE loss function and stochastic gradient descent with momentum $\rho=0.9$ with learning rate reduced whenever the network stops improving.

% \color{red!60}
% --potentially a figure showing the network stack--
% \color{black}

\section{Compliant motion control}

\begin{figure}[tb]
	\centering
	\includegraphics[width=6cm,keepaspectratio]{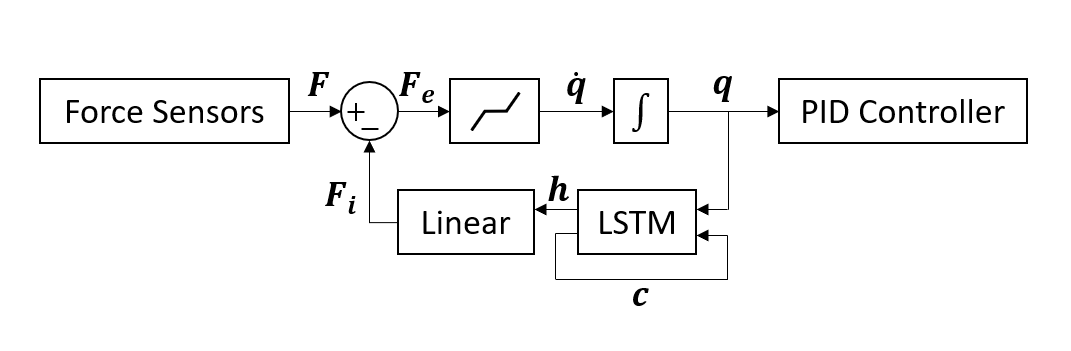}
	\caption{Diagram of the RNN-based compliant motion controller in the actuator space.}
	\label{fig:control_loop}
\end{figure}

Using the feed-forward tension estimator provided by RNN, an external force component can be extracted from sensor measurements and a compliant motion controller can be constructed to minimize this value, as shown in Fig. \ref{fig:control_loop}. 
Unlike typical controllers, which transform forces into the pose space, and determine the actuator space motion required to travel towards the desired conforming pose, we neglect to model the robot kinematics, and present an algorithm operating entirely in the actuator space. The cables have a very simple actuator-force relationship, namely pulling a cable increases tension, and releasing a cable decreases tension. Thus we take the unloaded tension prediction from the neural network and subtract it from the sensor measurements to obtain the external force,

\begin{equation}
\bm{F_{ext}} = \bm{F_{meas}} - \bm{F_{int}}.
\end{equation}

A deadband threshold $\lambda$ is required to prevent erroneous motion due to prediction and sensor inaccuracy, and forces exceeding the deadband produce a control velocity $\bm{\dot{q}}$ with a hand-tuned proportionality constant $\beta$.

\begin{equation}
\bm{\dot{q}} = 
    \begin{cases} 
        0 & -\lambda \leq \bm{F_{ext}}\leq \lambda \\
        -\beta(\bm{F_{ext}} - \lambda sign(\bm{F_{ext}})) & otherwise
    \end{cases}
\end{equation}

The threshold $\lambda$ is selected based on the performance of the neural network on the validation set at approximately $1.25\times$ mean error. The responsiveness constant $\beta$ is tuned to a suitable balance of sensitivity and stability. 
The performance of the controller will be evaluated in the next section.

\section{Experimental validation}
\subsection{Experimental setup}
Experiments were performed on a single segment platform with three braided steel wire tendons and a flexible backbone of reinforced plastic hose. 
Four tendon guide disks of radius 10~mm were spaced evenly along the length of the segment at 33~mm intervals. 
Cables were driven by standard metal gear servos with 13~kg$\cdot$cm of holding torque on a drive wheel of radius 20~mm providing 63~mm of actuation distance, and a maximum safe operating force of 65~N. 
Tension sensing was provided by strain gauges glued to small tabs of aluminium inserted between two segments of cable, routed through a custom amplification board to the analog-digital converters of a microcontroller unit. 
Measurements were taken at 20~kHz and provided to the control loop at 100~Hz after passing through a first-order low pass filter defined as
\begin{equation}
    y[n] = \frac{255}{256}y[n-1]+\frac{1}{256}x[n]
\end{equation}
where $x$ is the unfiltered tension reading and $y$ is the filtered value feeding the control loop.

Typically, the use of low-cost components, filtering and external PID controllers would incur significant additional modeling cost for control loop design.
However, the holistic neural network approach enables learning the combined system transfer function provided sufficient training data and input history.

\subsection{Tip force estimation}
As the incompressible backbone presents a curvy shape in the natural state, the majority of cable tension is balanced by the backbone and elastic energy, but differing cable tensions create a torque about the center of the tip segment producing a force in the plane of the tip segment. 
The ratio between cable tension, tip torque, and force in the tip plane is encapsulated in a couple constant $\alpha$. 
Projecting into the tip plane using the orientation of cable connections produces the form
\begin{equation}\label{eq:tip_t}
    \begin{bmatrix}
        F_x \\ F_y
    \end{bmatrix}
    =
    \frac{\alpha}{2}
    \begin{bmatrix}
        0 & -\frac{\sqrt{3}}{2} & \frac{\sqrt{3}}{2} \\
        1 & -\frac{1}{2} & -\frac{1}{2}
    \end{bmatrix}
    \begin{bmatrix}
        T_1 \\ T_2 \\ T_3
    \end{bmatrix}
\end{equation}
where the constant $\alpha$ represents the ratio between the force in a cable and magnitude of force in the tip acting in the corresponding direction. 

\begin{figure}[tb]
	\centering
	\includegraphics[height=5cm,keepaspectratio]{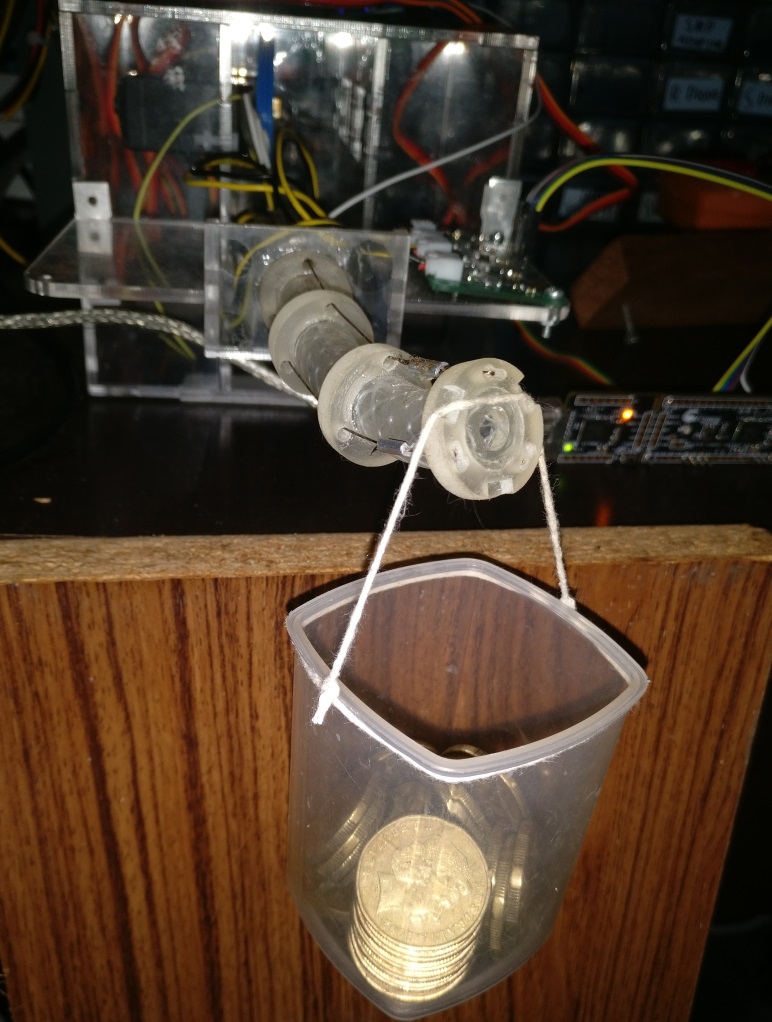}
	\caption{Experimental setup for tip-cable force calibration. Known weights (coins) were applied to the tip at various orientations and positions.}
	\label{fig:calib}
\end{figure}

\begin{figure}[tb]
	\centering
	\includegraphics[width=6cm,keepaspectratio]{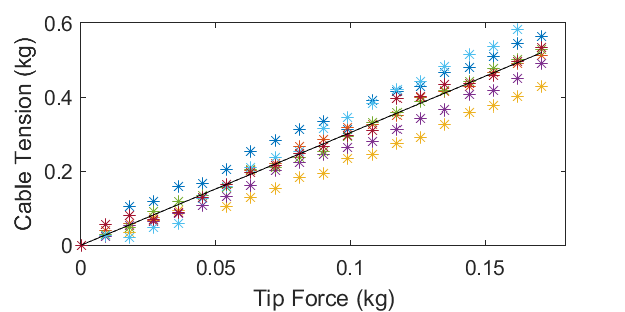}
	\caption{Calibration of the relationship between tip force and cable tension. The horizontal axis is the measured tip force and the vertical axis is the force calculated by the sensed external tension in each cable. Six trials of measurement were performed at each interval. The results show that $\alpha=1/3.04$ can be taken for the system. }
	\label{fig:external_calib}
\end{figure}

With the experimental setup shown in Fig. \ref{fig:calib}, the robot was moved to several different positions, and known weights were added to the tip in the form of 9 g dollar coins. 
The measured external tension in each cable was transformed into the tip plane using (\ref{eq:tip_t}) producing Fig. \ref{fig:external_calib}, from which we conclude that the constant $\alpha$ equals approximately 1/3 in this system.

Note this step is just for interpretation of the experimental results in the following sections.
It is not mandatory for the compliant motion control since the control is performed directly in the actuator space.

\subsection{Data collection for offline training}\label{sec:training}
%\color{red!60}--writing this, I realise that I would like to do a more general exploration algorithm, making collection of the training data also completely model-less, but alas, no time remains--\color{black}

Collection of training data for offline training requires moving the robot in such a way as to explore the input space with sufficient variation to accurately capture the underlying relationships. 
We used an elementary model to allow smooth traversal of the task space, transforming the position of the three cables $q_1,q_2,q_3$ to a 2D position $(x, y)$ and combined tension $c$ using the vectors from the tip center to the cable attachment points in the tip plane,

\begin{equation}
    \begin{bmatrix}
        x \\ y \\ c
    \end{bmatrix}
    =
    \begin{bmatrix}
        0 & -\frac{\sqrt{3}}{2} & \frac{\sqrt{3}}{2} \\
        1 & -\frac{1}{2} & -\frac{1}{2} \\
        1 & 1 & 1
    \end{bmatrix}
    \begin{bmatrix}
        q_1 \\ q_2 \\ q_3
    \end{bmatrix}.
\end{equation}

Moving a cable by one unit moves the position vector $(x,y)$ by one unit in the corresponding direction. 
The constraint that all cables must remain within a given tension range, to prevent damage to cables, or tension falling to 0 and invalidating sensor relationships was enforced by learning a surface
\begin{equation}
    c = f(x,y)
\end{equation}
such that the tension in each cable was within the desired range. 
The surface $f$ was explored by moving to new positions, and varying $c$ to achieve optimal sensing tension in each cable, and recording the point $(x, y, c)$ in a lookup table. 
Future movements sampled the surface using LOESS smoothing \cite{cleveland1988locally} and the exploration was continued until the suitable region of the $x,y$ plane was filled to sufficient fidelity. 
We note that the control model used for data collection does not need to be physically accurate, or complex, only topologically smooth, to allow reasoning about training motions.

In operation, we expect the robot to exhibit a range of motion styles, smooth curves, sudden changes in direction, and arbitrary starts and stops across a range of velocities. 
To collect representative training data, we created three parameters: velocity, jerkiness, and sleepiness. 
Velocity controls the rate at which we traverse the $(x,y)$ surface, jerkiness controls the maximum angular rate of change in $(x,y)$, and sleepiness causes the sudden cessation of movement for small lengths of time. 
Target positions were randomly selected from $(x,y)$ and changed whenever the robot entered a small radius of the current target. 
Every 20 seconds, a new set of velocity, jerkiness and sleepiness parameters were chosen.
The motor control signals and corresponding recorded cable tensions were recorded at 100~Hz for six hours producing 2~million points of training data for the RNN. 
We then used eighty per cent of the generated data for training, with the remaining 20\% kept for validation.

% \textcolor{red}{Zongyuan - consider to revise the following paragraph so as to align with Sec. V-D?}

% \begin{figure}[tb]
% 	\centering
% 	\includegraphics[width=8cm,keepaspectratio]{Figures/training_loss.png}
% 	\caption{Training and validation accuracy across training epochs}
% 	\label{fig:training_loss}
% \end{figure}

% Fig. \ref{fig:training_loss} \textcolor{red}{Zongyuan/David - where to place and refer to this figure?}
% \vspace{-0.5cm}
\subsection{Comparison of NN architectures}
In this section, we show the empirical performance of our deep sequential model LSTM with various hyper-parameter settings on the tension force prediction task. 
Furthermore, we also compared our RNN based LSTM model to a non-recurrent architecture convolutional neural network (CNN), which has outperformed LSTMs in neural machine translation~\cite{tang2018self}. 
CNN network is able to connect distant signals via shorter network paths than LSTMs. 
Although our task is very different from machine translation, it is worthwhile to test this theoretical argument empirically on another signal processing task and compare their differences. 

%\noindent\textbf{CNN:} 
CNNs are hierarchical based networks, and it captures local correlations in the signal.
Unlike RNNs, CNN is easier and faster to train due to its architectural design. 
The main difference between RNNs based models and CNNs is the ability to capture the long-range dependencies. 
The receptive range of a CNN is computed with respect to the number of layers $L$ and kernel size $k$. 
The longest context size is $L(k-1)$.
In this experiment we built a CNN with $L=3$ and kernel size $k=32,64$.
ReLU was attached to each convolution layer. 
Max-pool was employed on the feature map before connecting it to a fully-connected layer with 32 neurons. 

Recurrent and non-recurrent architectures are hard to compare fairly because many factors may affect the performance. 
We trained both models with similar hyper-parameters (e.g. input sequence length, batch size, number of hidden states in the fully-connected layer) and techniques in PyTorch~\cite{ketkar2017introduction} except for those parameters specific with each architecture (e.g. number of epochs to converge).\footnote{The RNN based network was trained for 2000 epochs in 16~hours on a GTX 1070 with batches of 20~k points, While CNN based network could converge within 100 epochs in less than 3 hours training time.}

Table~\ref{table:lstm/cnn} gives the performance of all the architectures, including experiments with different input sequential length and number of neurons or filters in the hidden layer. 
The evaluation metric being used here is mean error. 
Recurrent models LSTM distinctly outperform CNN models, showing the importance of capturing long-term dependency in force prediction. 
Moreover, the majority of LSTM and CNNs achieve lower validation error when given more hidden states. 
We attribute this observation to the model capacity, which makes the model fit better to the training dataset. 
For both models, sequential length plays an important role in capturing long-term dependency. 
This becomes more obvious when comparing the performance gain between \textit{LSTM-64} (-0.42) and \textit{CNN-64} (-0.34). 
This affirms our assumption that recurrent models are better long-term features extractors than CNNs, and thus are more suitable for the task discussed in this paper.          

\begin{table}[!t]
  \centering
  \caption{\small{The results of different LSTM and CNN models on force prediction task.}}
  \vspace*{0.2cm}  
  \begin{tabular}{|c|c|c|}
    \hline
    \bf Model & \bf Sequential length  & \bf Mean err (kg)    \\ \hline
    \textit{LSTM-32}    & 100/200  & 0.304/0.282  \\ \hline
    \textit{LSTM-64}    & 100/200  & 0.300/\textbf{0.258}   \\ \hline
    \textit{CNN-32}                & 100/200  & 0.287/0.295   \\ \hline
    \textit{CNN-64}                & 100/200  & 0.304/0.270   \\ \hline
  \end{tabular}
  \label{table:lstm/cnn}
\end{table}

% \color{red!60}I'm not sure exactly what the best way to present this is, and what portions of the full training table to show.\color{black}

% \begin{itemize}
%     \item Sequence length, prediction accuracy vs length. Increases monotonically for all models from 1,5,20,50,100,200
%     \item LSTM 64 \textgreater Sigmoid 256 \textgreater LSTM 32 \textgreater Sigmoid 126 \textgreater DCNN 32
%     \item (note, DCNN 64 not comparable due to training memory limits, would've done 50 if enough time)
%     \item LSTM and DCNN have equally low overfit \textless Sigmoid
%     \item Sigmoid and DCNN both fast to train \textless LSTM (by about 4x)
%     \item Second (more generalised dataset) reduced overfit across all models. LSTM still \textgreater DCNN but not by much. Sigmoid better by far for both accuracy and training time when overfit is no concern.
%     \item A dropout layer before the final Linear layer did not help at all
%     \item 3x200 -\textgreater Encoder (various) -\textgreater feature vec -\textgreater Linear (decoder) -\textgreater 3x1
% \end{itemize}

\subsection{Responsiveness evaluation}

\begin{figure*}[tb]
	\centering
	\includegraphics[width=16cm,keepaspectratio]{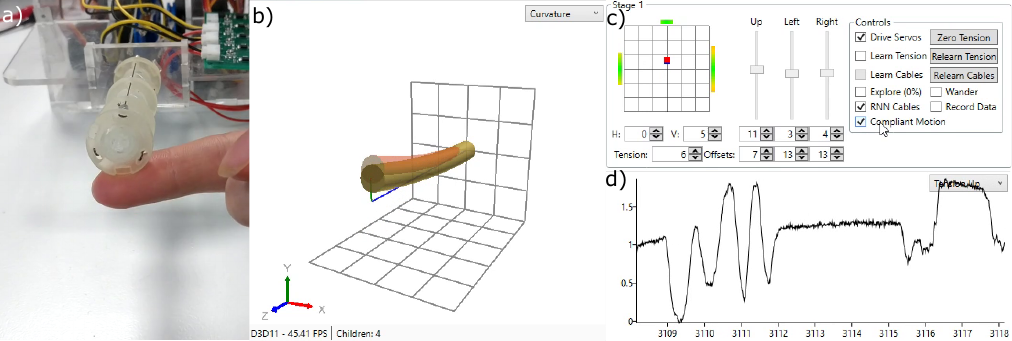}
	\caption{Responsiveness evaluation. a) A force was applied to the tip of the robot. b) Estimation of the RNN. Yellow and brown parts represent the current state and estimated deformation based on cable tension measurement, respectively. c) Control panel. d) Cable tension measurement.}
	\label{fig:response}
\end{figure*}
With the RNN-based tension estimator validated in the last section, a compliant motion controller was implemented in the actuator space.
All elements of the control loop, sensor measurements and the neural network were discretised at 100~Hz. 
Control was implemented in C\# on the windows platform with accompanying UI, communicating with a slave microcontroller which supplied sensor readings and motor control input over low-latency (3~ms) serial. 

\begin{figure}[tb]
	\centering
	\includegraphics[width=6cm,keepaspectratio]{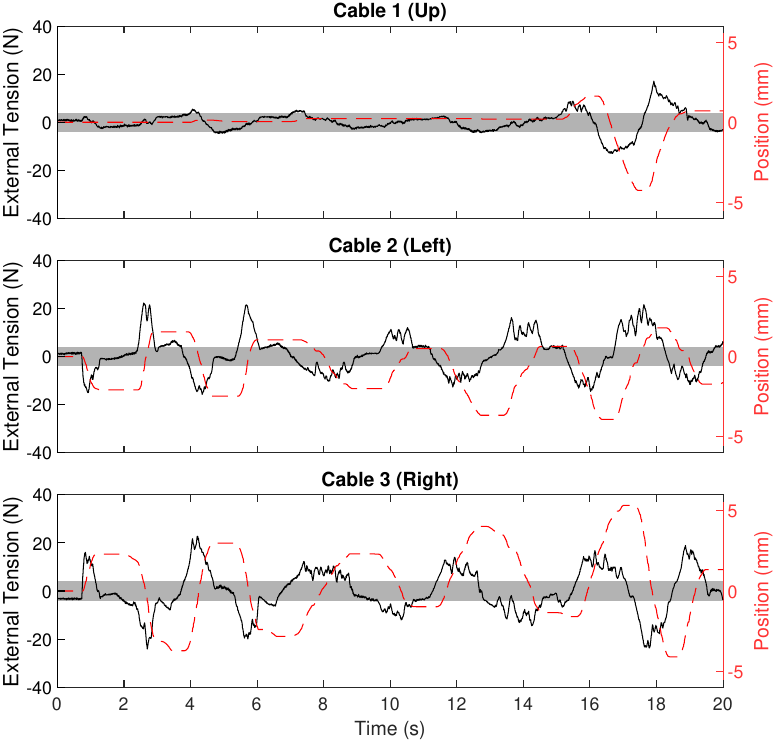}
	\caption{Compliant response to measured external forces throughout various motions with deadband (gray).}
	\label{fig:compliant_response}
\end{figure}

The responsiveness of the controller was first evaluated by manually applying a force to the tip, as shown in Fig. \ref{fig:response}.
The RNN was able to predict the internal tension in each cable and thus the external tension was separated to drive the actuators.
Fig. \ref{fig:compliant_response} shows the control system performance compliantly in response to forces applied to the tip of the robot. 
When the measured external force on a cable exceeded the deadband threshold, the cable moved rapidly to reduce the input force at a rate proportional to the magnitude of the input.  
%\color{red!60}--multimedia extension (youtube video 1)--\color{black}

\begin{figure}[tb]
	\centering
	\includegraphics[width=6cm,keepaspectratio]{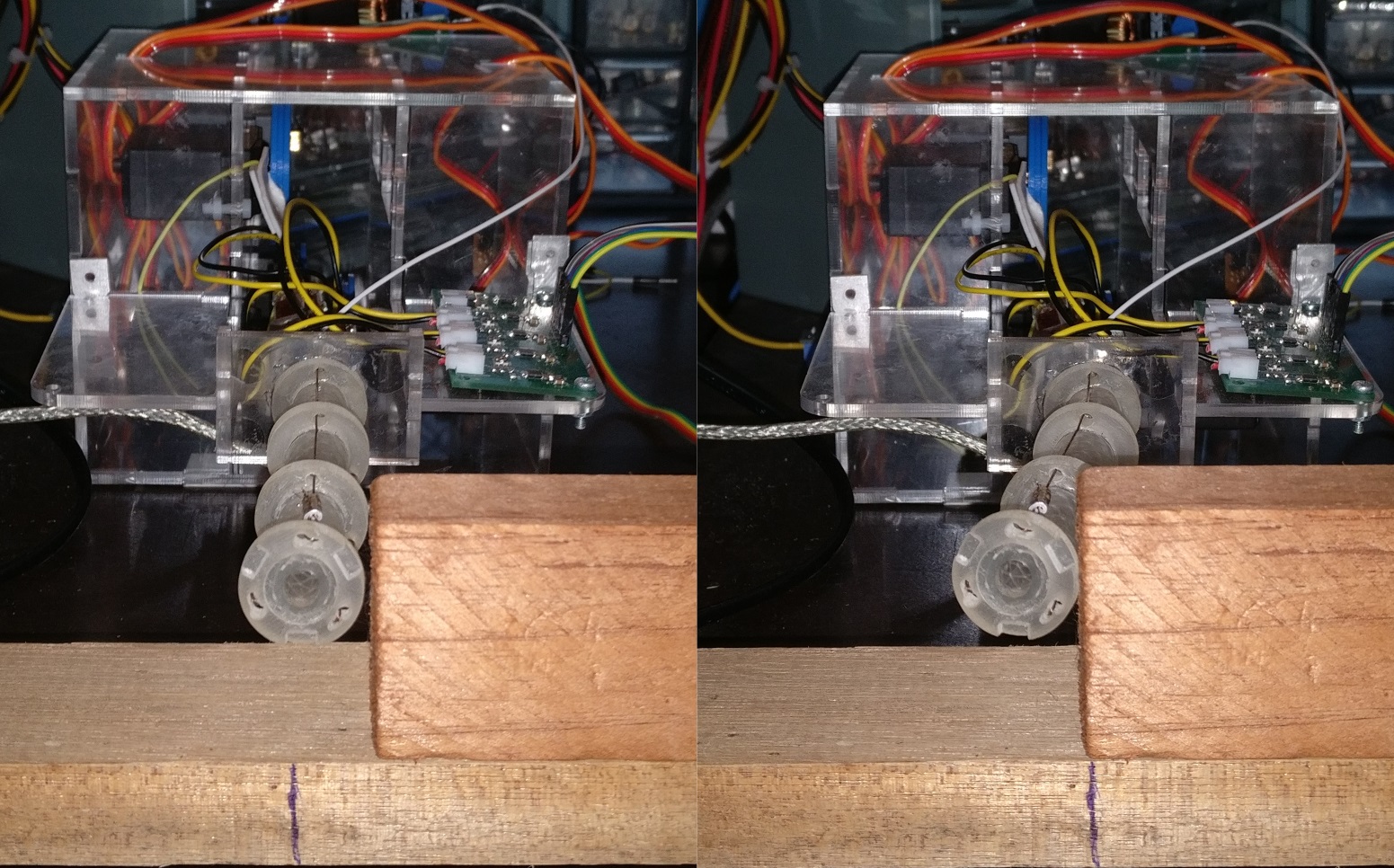}
	\caption{Experimental setup for impulse response. The block is moved to the black line, applying a deflection to the robot tip.}
	\label{fig:impulse}
\end{figure}

\begin{figure}[tb]
	\centering
	\includegraphics[width=6cm,keepaspectratio]{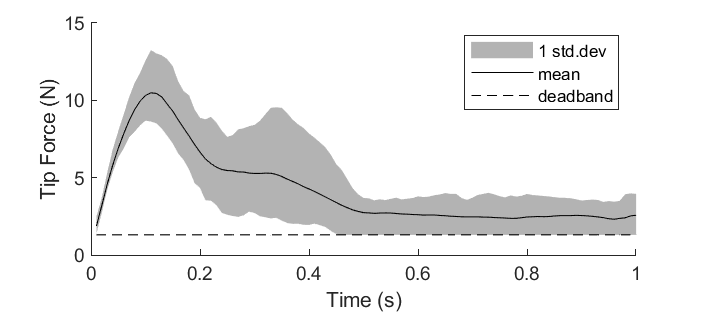}
	\caption{Tip impulse reduction with compliance.}
	\label{fig:impulse_response}
\end{figure}

In order to further evaluate the response time and sensitivity of the controller, a wooden block was used to push the tip backwards and forwards (Fig. \ref{fig:impulse}), stopping at a marked line and waiting for the controller to reduce the interaction force. Fig. \ref{fig:impulse_response} shows the mean tip force in response to 20 impulses of varying intensity. Individual responses were correlated using the moment the force on one cable exceeded the dead-band threshold. The controller rapidly moved to reduce the impulse in the first 0.2~s, and then approached the dead-band asymptotically. 


\subsection{Compliant insertion}
\begin{figure}[tb]
	\centering
	\includegraphics[width=6cm,keepaspectratio]{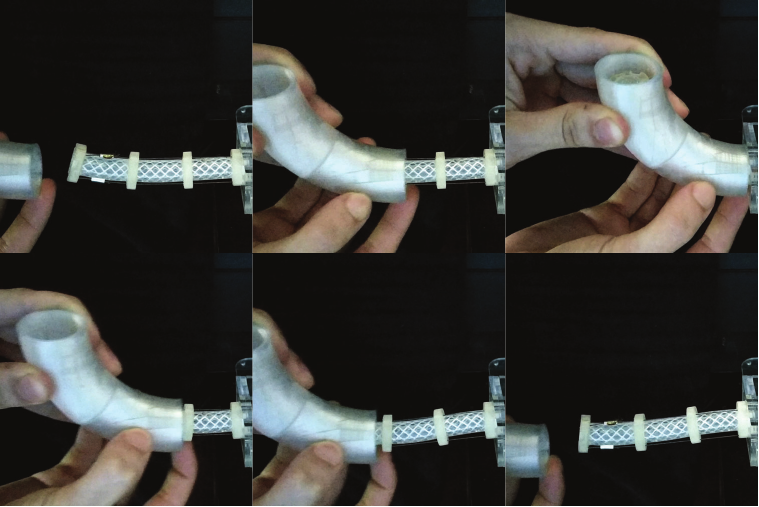}
	\caption{Individual frames of compliant insertion and retraction demonstration. }
	\label{fig:insertion_frames}
\end{figure}

To evaluate the controller performance when exploring unknown environments, the robot was repeatedly inserted into and retracted from a curved tube of radius 13 mm. 
Illustrated in Fig. \ref{fig:insertion_frames}, the robot successfully conformed to the tube shape, achieving a 90 degree bend with insertion, and returning to a straight shape during retraction. 
The controller had no knowledge of tube shape and conforms solely in response to contact between the robot and tube surface. 

\begin{figure}[tb]
	\centering
	\includegraphics[width=8cm,keepaspectratio]{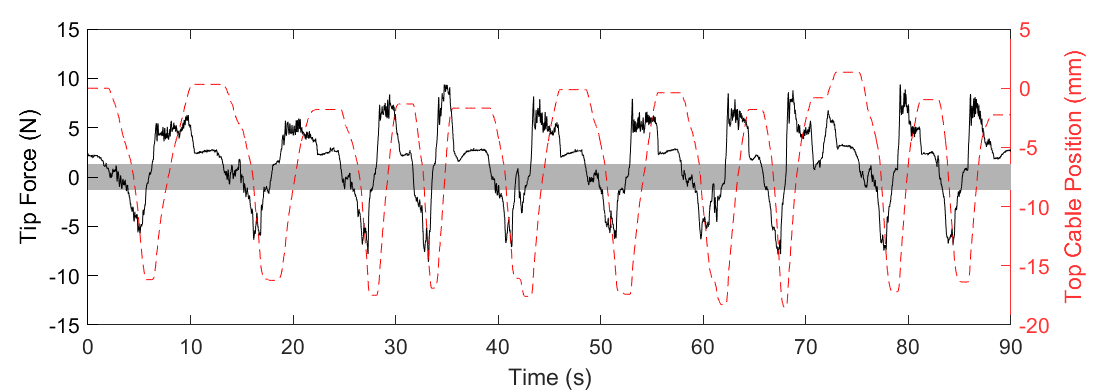}
	\caption{Reaction to external forces during repeated insertion trials with deadband (grey).}
	\label{fig:insertion_response}
\end{figure}

\begin{figure}[tb]
	\centering
	\includegraphics[width=7cm,keepaspectratio]{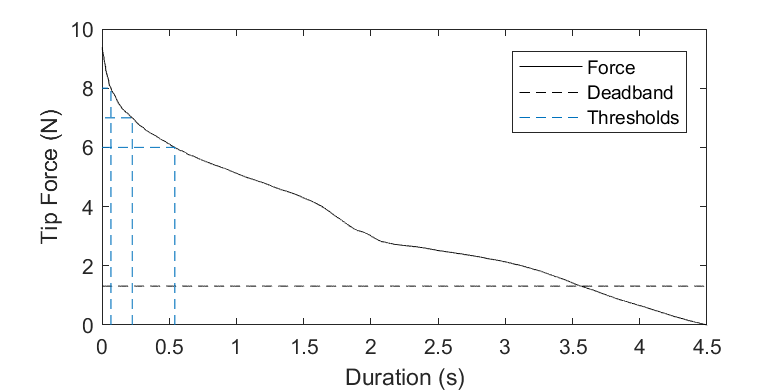}
	\caption{Duration for which tip forces exceeding a given threshold are experienced in the average insertion.}
	\label{fig:insertion_forces}
\end{figure}

Ten insertions were performed over a 90~s period, with reaction data shown in Fig. \ref{fig:insertion_response}. 
During the insertion, contact force was maintained at 5-7~N with larger forces reduced more quickly providing a faster insertion. 
Fig. \ref{fig:insertion_forces} shows a cumulative histogram indicating the duration for which forces exceeding a given threshold were experienced, averaged over insertion trials. 
Tip forces exceeding 6~N, 7~N and 8~N were experienced for an average duration of 0.54~s, 0.23~s and 0.07~s respectively. 
The average insertion/retraction was completed in 3.5~s at a speed of 29~mm/s and cable movement speed of 5~mm/s.
These results show that the developed controller based on RNN was capable of endowing the continuum robot with active compliance while exploring unknown environments.
%\color{red!60} --this describes the results, but does not analyse or compare them--\color{black}

\section{Discussion}
The experiments show that a complex model of the mechanics of the continuum robot is not necessary for active compliant motion control.
However, this does not mean having a model of the robot is useless.
A model, even if being inaccurate, can direct the motion of the robot in the data collection procedure for offline training.
Moreover, the complex model is avoided by implementing the controller in the actuator space directly for the compliant motion control task.
For tasks where the tip of the robot needs to apply a certain force, an accurate model relating the external force output and the measurements of the actuating cables may be needed.
This could also be achieved using machine learning techniques given a considerable amount of calibration data.
However, the performance of such an approach is yet to be evaluated.

%\color{red!60}
%For compliance, tip forces are not required.
%For tasks like palpation, force estimation is required. In such cases, a model to relate the cable tension to the tip force may be needed. Alternatively, such a model may be learned as well when measurements of the tip forces are available for training.
% \color{black}

\section{Conclusion}
Compliance is one of the key capabilities of continuum robots to enhance safety while used in minimally invasive surgery.
Active compliant motion control has been developed based on complex derivation of the mechanics model of continuum robots.
This paper proposed a model-less approach leveraging an RNN to capture the highly nonlinear factors of the system such as hysteresis of the cables, friction, and delays of electronics, that are usually difficult to model accurately.

Using a 3-tendon single-segment continuum robot with force sensors on each cable, an RNN was trained to predict the internal cable forces produced by backbone stiffness, friction and other factors from the actuator control signals during unloaded motion. 
Then, a feed-forward motion controller was developed directly in the actuator space to minimize the difference between the predicted cable tension provided by the RNN and the measured tension.
The superiority of the RNN over the other architectures was validated by comparison.
A set of experiments were conducted to demonstrate that the continuum robot with an RNN based feed-forward controller was capable of responding to external forces quickly and entering an unknown environment compliantly.

While the positive results on the simple platform show promising feasibility of the machine learning based approach for constructing compliance controller, more experiments need to be carried out on continuum robots that are better designed for real applications.
In addition, an extension of the method to multiple-segment continuum robots is worth further study.

\bibliographystyle{IEEEtran} 
\bibliography{bib}

\end{document}